\documentclass[letterpaper]{article} 
\usepackage{aaai23}  
\usepackage{times}  
\usepackage{helvet}  
\usepackage{courier}  
\usepackage[hyphens]{url}  
\usepackage{graphicx} 
\usepackage{natbib}  
\usepackage{caption} 
\usepackage{multirow}
\usepackage{comment}
\usepackage{amsmath}
\usepackage{amsfonts}
\usepackage{amssymb}
\usepackage{amsbsy}
\usepackage{xcolor}
\usepackage{mathtools}
\usepackage{algorithm}
\usepackage{algorithmic}

\usepackage{newfloat}
\usepackage{listings}
\DeclareCaptionStyle{ruled}{labelfont=normalfont,labelsep=colon,strut=off} 
\floatstyle{ruled}
\newfloat{listing}{tb}{lst}{}
\floatname{listing}{Listing}
\title{Newell's theory based feature transformations for spatio-temporal traffic prediction}
\author{
    Agnimitra Sengupta\textsuperscript{\rm 1}, 
    S. Ilgin Guler\textsuperscript{\rm 2}\equalcontrib\\
}
\affiliations{
    \textsuperscript{\rm 1,2}Department of Civil \& Environmental Engineering, The Pennsylvania State University\\
    University Park, PA 16802, USA\\
    \textsuperscript{\rm 1}avs6897@psu.edu, \textsuperscript{\rm 2}sig123@psu.edu
}

\usepackage{bibentry}

\begin{document}

\maketitle
\begin{abstract}
Deep learning (DL) models for spatio-temporal traffic flow forecasting employ convolutional or graph-convolutional filters along with recurrent neural networks to capture spatial and temporal dependencies in traffic data. These models, such as CNN-LSTM, utilize traffic flows from neighboring detector stations to predict flows at a specific location of interest. However, these models are limited in their ability to capture the broader dynamics of the traffic system, as they primarily learn features specific to the detector configuration and traffic characteristics at the target location. Hence, the transferability of these models to different locations becomes challenging, particularly when data is unavailable at the new location for model training. To address this limitation, we propose a traffic flow physics-based feature transformation for spatio-temporal DL models. This transformation incorporates Newell's uncongested and congested-state estimators of traffic flows at the target locations, enabling the models to learn broader dynamics of the system. Our methodology is empirically validated using traffic data from two different locations. The results demonstrate that the proposed feature transformation improves the models' performance in predicting traffic flows over different prediction horizons, as indicated by better goodness-of-fit statistics. An important advantage of our framework is its ability to be transferred to new locations where data is unavailable. This is achieved by appropriately accounting for spatial dependencies based on station distances and various traffic parameters. In contrast, regular DL models are not easily transferable as their inputs remain fixed. It should be noted that due to data limitations, we were unable to perform spatial sensitivity analysis, which calls for further research using simulated data.
\end{abstract}

\section{Introduction}
Short-term traffic forecasts play a crucial role in supporting operational network models and facilitating Intelligent Transportation Systems (ITS) applications. Researchers have long recognized the importance of accurately predicting future traffic conditions for proactive traffic management \cite{cheslow1992} and comprehensive traveler information services \cite{kaysi1993}. 
These forecasts enable ITS applications to provide drivers with precise and timely information, allowing them in making informed decisions. 
As a result, this enables effective congestion mitigation, reduced travel time, and an enhanced overall travel experience. The success of these strategies relies heavily on the quality and accuracy of the traffic forecasts, which are obtained through sophisticated time-series forecasting models.

The prediction of traffic conditions, including flow, occupancy, and travel speed, is essentially a time-series forecasting problem that relies on input data from a sufficient number of spatially distributed sensors throughout the network. Therefore, in addition to considering temporal dependencies, it is crucial to incorporate the spatial correlations among various sensor data within the network to effectively capture the intricate dynamics of traffic flow. 

Various parametric techniques have been used to model temporal dependence in traffic time series, including historical average algorithms, smoothing techniques, and autoregressive integrated moving average (ARIMA) models \cite{ahmed1979,levin1980}. However, studies have indicated that ARIMA models have limitations in capturing complex traffic patterns, especially during congested conditions \cite{davis1991,hamed1995}. Additionally, the development of multi-variable prediction models incorporating flow, speed, and occupancy for traffic forecasting has yielded mixed results \cite{innamaa2000,dougherty1997,florio1996,lyons1996}. These findings highlight the need for more advanced and effective modeling approaches in traffic prediction.

Alternately, state-space models have been shown to be an excellent foundation for modeling traffic data since they are multivariate in nature and can describe simpler univariate time series \cite{okutani1984, chen2001, chien2003, whittaker1997}. Following advances in computer efficiency and capacity to handle large data quantities, non-parametric techniques, which, unlike parametric methods, do not specify any functional form, was used to model traffic data with greater transferability and robustness across datasets \cite{smith1997, clark2003}. 
These approaches, in contrast to parametric methods, have more degrees of freedom, allowing them to better adapt to non-linearities and capture spatio-temporal features of traffic. For small scale traffic scenarios, methods like nearest neighbors \cite{smith2002}, support vector machine \cite{mingheng2013}, and Bayesian network \cite{sun2006} have proven to be beneficial. However, these methods become limited in their ability to extract effective features for large-scale traffic analysis \cite{lin2019}. 

To the contrary, neural networks (NN) and deep learning (DL) can use a multi-layer architecture to capture complex relations \cite{lecun2015} in traffic data. Use of NNs to capture temporal patterns in traffic time series have been demonstrated in \cite{chang1995,innamaa2000,dia2001}. In this backdrop, numerous variants of NN were developed, such as backpropagation neural networks \cite{park1999}, the modular neural network \cite{park1998} and radial basis function neural network \cite{parkcarroll1998}. However, for sequential data like traffic time series, recurrent neural networks (RNN) \cite{van2002,rumelhart1986} and its variants like long short-term memory (LSTM) \cite{hochreiter1997} have been specifically introduced to preserve temporal correlations between observations to predict future states. For example, Ma et al. \cite{ma2015} used an LSTM architecture to make short-term travel speed predictions. Yu et al. \cite{yu2017} applied LSTM and autoencoder to capture the sequential dependency for predicting traffic under extreme conditions, particularly for peak-hour and post-accident scenarios. Cui et al. \cite{cui2020} proposed a deep stacked bidirectional and unidirectional LSTM network for traffic speed prediction.

To account for the spatial aspects of traffic, data from multiple locations across the network need to be utilized in predicting future traffic states. For instance, Stathopoulos and Karlaftis \cite{stathopoulos2003} incorporated data from upstream detectors to improve predictions at downstream locations using a multivariate time-series state-space model. DL models like deep belief networks \cite{huang2014} and stacked autoencoder \cite{Lv2014} have also been considered for multi-sensor information fusion - however, because of the underlying structure of the input data, the spatial relationship could not be efficiently captured with these models. To this end, convolutional neural network (CNN) \cite{krizhevsky2012} has emerged as one of the most successful deep NNs to model topological locality or spatial correlation by use of filters or kernels to extract local features. For example, \cite{zhang2017} developed a DL model, ST-ResNet which used convolutions for city-wide crowd flow prediction. However, CNNs are particularly successful when dealing with data in which there is an underlying Euclidean structure. For generalized, non-Euclidean data, graph convolutional neural network (GCNN) \cite{henaff2015} have been used for traffic network modeling and prediction tasks \cite{cui2019,mallick2021}. 
However, while these studies explicitly model temporal or spatial dependency, they do account for their interaction effects. Recent studies proposed integrating CNN and LSTM \cite{yao2019,yao2018} to jointly model spatial and temporal patterns in traffic. Several other hybrid models like Sparse Autoencoder LSTM \cite{lin2019}, DeepTransport \cite{cheng2018} - hybrid CNN and RNN with an attention mechanism, ST-TrafficNet \cite{lu2020} have been recently demonstrated to efficiently capture spatio-temporal features of traffic data. 

In modelling traffic data, DL algorithms have shown considerable potential. The success of data-driven models can be linked to the data quality and quantity. However, limited data availability poses a significant restriction to model development for many freeway corridors. Transfer learning (TL) is a promising method for avoiding data scarcity, training, and deployment issues. In this method, a model learned for one task is reused and/or altered for a similar task. TL is commonly employed for image classification, sentiment analysis, and document classification \cite{pan2009,zhuang2020}, but it has received less attention in the traffic forecasting domain\cite{mallick2021}. Further, physics-informed models may uncover additional dynamics of the system, that might not be observable from the limited data. Physics-informed deep learning (PIDL) models usually comprise of a model-driven component (a physics-informed NN for regularization) and a data-driven component (a NN for estimation) to integrate the advantages of both components. Using this framework, recent researches \cite{Huang2020physics, shi2021physics,thodi2022physics} have demonstrated the superiority of the PIDL in traffic state estimations over purely data-driven or traffic flow physics-based approaches. This paper presents yet another approach to incorporate traffic flow physics to a DL model that uses a physics-based feature enhancement to perform the task of traffic flow prediction along a freeway corridor. Our approach is particularly advantageous due to its ability to be transferred to new locations where data is not available. 

The remainder of the paper is organised as follows: first, we present the conceptual background of Newell's solution to Lighthill-Whitham-Richards (LWR) model and the DL model, followed by the physics-based feature enhancement to the model. Next, the problem setup is described, including the modelling results. Finally, some concluding remarks are presented.

\section{Background}
In this section, we provide a brief introduction to Newell's simplified solution to the Lighthill-Whitham-Richards (LWR) model - which forms the basis of the physics-based feature transformation, followed by an overview of the DL model considered.

\subsection{Newell's simplified theory estimations} \label{section:newell}
Traffic state evolution on a roadway using the basic principle of LWR continuum model requires identification of `shockwaves' and their interactions on the time-space diagram. A simplified approach - Newell's solution \cite{newell1993} - can be used to estimate traffic dynamics at a particular location in terms of cumulative vehicle counts, i.e., the cumulative number of vehicles observed at that location. Using vehicle counts at a detector location, this method aims to estimate cumulative vehicle counts at some other location along the homogeneous freeway assumed to exhibit a triangular fundamental diagram. Cumulative vehicle counts at detector stations can be appropriately translated using fundamental traffic parameters (defined in Table~\ref{tab:notations}) to give an estimate of cumulative counts at the target site consistent with traffic flow theory. The relevant components of Newell's solution are summarized below. The basic idea is to determine the change in vehicle counts along characteristics of free-flow from upstream and congested flow from downstream. Therefore, this approach \textit{assumes upstream is in free-flow state, whereas the downstream is in congestion}. Newell's simplified theory states that the real state at the target site would be determined as the minimum of what would be expected along the free-flow characteristic and the congested characteristic.

\begin{table}[!ht]
	\caption{Macroscopic fundamental variables}\label{tab:notations}
	\begin{center}
\begin{tabular}{ll} 
\hline
Symbols & Variable \\ \hline
$v_f$ & Free flow speed \\
$w$ & Congested wave speed \\
$N_i$ & Vehicle count at station i \\
$q$ & Flow \\
$q_c$ & Flow at capacity\\
$k$ & Density \\
$k_c$ & Density at capacity\\
$k_j$ & Jam Density\\ \hline

\end{tabular}
\end{center}
\end{table}

\subsubsection{Along the free-flow characteristic}
Assume a point $A$ in a free-flow state (F) with vehicle count $N_A$. Consider a target location $B$ downstream to location $A$. As we move from $A$ to $B$, which lies on the characteristic that emanates from $A$, the change in vehicle counts is given by $\Delta N_{A\rightarrow B}$. 

\begin{equation}
\begin{split}
    \Delta N_{A\rightarrow B} & = \dfrac{\partial N}{\partial x}\Delta x + \dfrac{\partial N}{\partial t}\Delta t \\
    \dfrac{\Delta N_{A\rightarrow B}}{\Delta x} & = \dfrac{\partial N}{\partial x} + \dfrac{\partial N}{\partial t}\dfrac{\Delta t }{\Delta x} \\
    & = -k + \dfrac{q}{v_f} \\
    \Delta N_{A\rightarrow B} & = 0\hspace{2mm}(\text{since}\hspace{2mm}q=kv)
\end{split} \label{eqn:nab}
\end{equation}

From Equation~\ref{eqn:nab}, the change in vehicular count along the interface or signal travelling at free flow speed is zero. However, the time for the signal to travel from $A$ to $B$ is $d_{A\rightarrow B}/v_f$, where $d_{A\rightarrow B}$ is the distance between locations $A$ and $B$. Therefore, the counts at $B$ as a function of counts at $A$ can be given by:

\begin{equation}
    N_f(t,B)=N\left( t-\dfrac{d_{A\rightarrow B}}{v_f}, A \right)\label{eqn:ff1}
\end{equation}

Conversely, if the location $B$ is upstream to location $A$, and free-flow regime persists, then Equation~\ref{eqn:ff1} can be suitably modified as,
\begin{equation}
    N_f(t,B)=N\left(t+\dfrac{d_{A\rightarrow B}}{v_f}, A \right)\label{eqn:ff2}
\end{equation}

\subsubsection{Along the congested flow characteristic}
Consider a target location $Y$ upstream to location $X$, that has a known congested state (C) with vehicle counts $N_X$. As we move from $X$ to $Y$ that lies on the characteristic that emanates from $X$, the change in vehicle counts is given by $\Delta N_{X\rightarrow Y}$.

\begin{equation}
\begin{split}
    \Delta N_{X\rightarrow Y} & = \dfrac{\partial N}{\partial x}\Delta x + \dfrac{\partial N}{\partial t}\Delta t \\
    \dfrac{\Delta N_{X\rightarrow Y}}{\Delta x} & = \dfrac{\partial N}{\partial x} + \dfrac{\partial N}{\partial t}\dfrac{\Delta t }{\Delta x} \\
    & = -k - \dfrac{q}{w} \\
    \Delta N_{X\rightarrow Y} & = -\Delta x\cdot(k_j) \\ & = d_{X\rightarrow Y}(k_j)
\end{split}\label{eqn:nxy}
\end{equation}

Therefore, in case of congestion, there is a finite change in the vehicle counts equal to $d_{X\rightarrow Y} k_j$ that occurs in time $d_{X\rightarrow Y}/w$. Thus, the counts at $Y$ as a function of counts at $X$ is given by:

\begin{equation}
    N_c(t,Y)=N\left( t-\dfrac{d_{X\rightarrow Y}}{w}, X \right) + d_{X\rightarrow Y} k_j \label{eqn:cc}
\end{equation}

The free-flow and congested estimators of cumulative counts at a target location using cumulative counts from another location along the homogeneous roadway are represented by Equations~\ref{eqn:ff1}, \ref{eqn:ff2} and \ref{eqn:cc}, which we suitably utilize in designing the physics-based model. 

\subsection{Spatio-temporal model}

In this study, we employ a combination of convolutional neural network (CNN) \cite{krizhevsky2012} and long short-term memory (LSTM) \cite{hochreiter1997}, referred to as CNN-LSTM, as the baseline model for jointly capturing the spatio-temporal aspects of traffic states.

CNN and LSTM extract information from the input data from two different perspectives - learning the time-invariant spatial characteristics in CNN; and short- and long-term temporal patterns in LSTM. 
Specifically, the convolutional layer in the model helps learn an internal representation of a two (or higher)-dimensional input through a feature learning process using kernel functions or filters. Notably, the convolution allows the model to learn features that are invariant across the time dimension. These features are then passed through an activation function to introduce non-linearities into the mapping function. Other layers, such as pooling layer are also used to reduce the number of parameters, while dropout layer prevents the model from overfitting. As a result, it generates more accurate feature representations, allowing the LSTM layers to learn temporal patterns with greater accuracy. 
LSTM uses feedback mechanism and several gating mechanisms where output from a previous time step is fed as input to the current step such that selective information from the past can propagate into the future states -- allowing them to persist, which makes it suitable for capturing the temporal evolution of traffic states.

\begin{figure}[!htb]
  \centering
  \includegraphics[width=0.8\columnwidth]{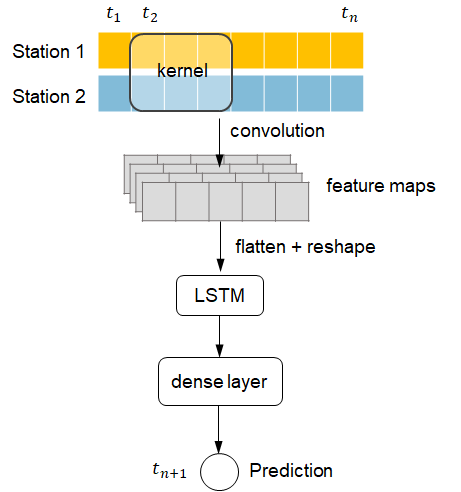}
  \caption{Architecture of CNN-LSTM model}\label{fig:model}
\end{figure}

A schematic of the model architecture considered in the study is shown in Figure~\ref{fig:model}. The architecture comprises a convolutional layer that processes the spatio-temporal input to extract relevant features using filters. The input to the model is structured as an $N \times n$ array that is constructed using data from $N$ fixed-location stations, with each row representing flows from a particular station over a period of time, $n$.
Notably, the pooling layers were not utilized since the spatial dimension of traffic data is limited. After the convolutional layer, a flattening layer is employed to reshape the outputs appropriately. The resulting feature outputs from the convolutional operation are then fed into an LSTM, which generates outputs through a sequence of densely connected layers.

\section{Methodology}
\subsection{Feature transformation} 
DL models learn the spatial and temporal characteristics specific to a detector configuration for which it has been trained using the traffic flows from neighboring locations. In other words, it is trained to map temporal patterns from the given spatial configuration to its target location. However, this mapping is specific to the exact location of detectors from which data is used (e.g., the distance between input and target detectors). Therefore, these models are difficult to adapt to varied configurations and hence cannot perform suitably, i.e. they are not transferable. To improve the transferability of the models, we aim to utilize knowledge from traffic flow theory and propose a physics-based feature transformation to the model inputs. Notably, we explicitly assume that the data available at the transfer location is limited and hence, cannot be used to train a new model. 

In our proposed approach, instead of using the flow data from the input detectors, the inputs are modified using Newell's transformations. The specific modification depends on whether the target location is upstream or downstream of the location from which input data is utilized. Hence, it is expected to learn generalized features independent of the distance between the detectors and hence be suitable for transfer learning. 

The physics-based modified version in this study includes three approaches: `Physics FC', `Physics FF' and `Hybrid'. In all three approaches, it is assumed that free-flow conditions prevail at an upstream station (e.g., location $A$) and the cumulative counts for the target or transfer station ($X$) are estimated using Equation~\ref{eqn:ff1} based on the cumulative counts at station $A$. The three approaches differ in how they treat input from a downstream station $B$. The `Physics FC' approach assumes that congestion prevails at a station downstream of the target or transfer stations, and hence uses Equation~\ref{eqn:cc} on the cumulative counts at station $B$ to determine the cumulative counts for the target or transfer station, $X$. To the contrary, the `Physics FF' approach assumes that free-flow prevails at the downstream station, and hence uses Equation~\ref{eqn:ff2} on the cumulative counts at station $B$ to determine the cumulative counts for the target or transfer station, $X$. 
The `Hybrid' approach, on the other hand, does not make explicit assumptions regarding the prevailing conditions (free-flow or congestion) at the station downstream of the target or transfer locations. Instead, it considers two inputs for the target or transfer station, $X$ -- one calculated using Equation~\ref{eqn:ff2} based on the cumulative counts at station $B$, and the other calculated using Equation~\ref{eqn:cc} based on the cumulative counts at station $B$. This approach utilizes both pieces of information to estimate the cumulative counts for the target or transfer station, thereby combining aspects of both free-flow and congestion conditions. Once the cumulative counts are computed for each case, the corresponding flows are then calculated by taking the differences between consecutive observations of the cumulative counts.

It is important to note that accurate estimation of fundamental traffic parameters such as free-flow speed ($v_f$), congestion wave speed ($w$), and jam density ($k_j$) is crucial for the proper functioning of these transformations. The estimation of these parameters will be discussed in the following section.





\subsection{Estimation of traffic parameters using detector data}
In our methodology, we demonstrate the estimation of traffic parameters using data from multiple loop detectors. For a stretch of freeway, we combine data from multiple detectors to obtain mean estimates of these parameters. This aggregation of detector data assumes homogeneity within the freeway section, implying that the traffic parameters do not significantly vary within the section. This assumption aligns with the assumptions made in Newell's solutions. However, it is important to note that for a larger number of detectors with wider spatial coverage, this assumption may not hold true.

\begin{figure}[!htb]
\centering
\includegraphics[width=0.90\columnwidth]{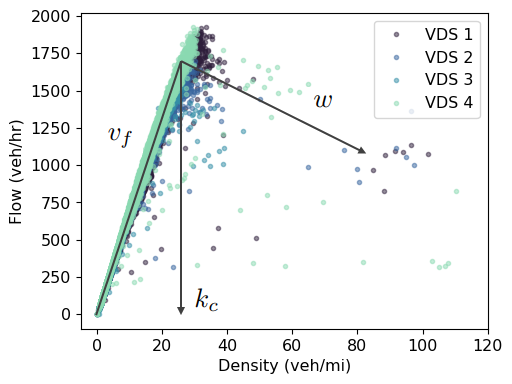}
\caption{Flow-density relationship between VDS 401826 and VDS 414041 on SR04 California}\label{fig:FD}
\end{figure}

The data provided includes measurements of flow ($q$), occupancy ($o$), and speed ($v$) recorded at 5-minute intervals from each station. Assuming a triangular fundamental diagram, we use the fundamental equation of traffic, $q = k \times v$, to calculate the corresponding densities ($k$). To determine the capacity ($q_c$) and free-flow speed ($v_f$), we consider the $95^{th}$ percentile values of flow and speed, respectively. This is done because the maximum flow and speed values do not indicate a stable state and persist for only a short period of time. The critical density ($k_c$) is then evaluated using the fundamental equation of traffic as $k_c = q_c/v_f$. These parameters provide essential insights into the traffic conditions and are fundamental for our subsequent analysis. Figure~\ref{fig:FD} illustrates the flow versus density relationship for four stations along the SR04 California highway.

The freeway operates in free-flow conditions for most of the day with congested states appearing only during peak hours. Therefore, congested states ($k>k_c$) are not easily observable in the flow-density curve, which do not allow us to directly estimate the full congested branch of the fundamental diagram, including congestion wave speed ($w$) and jam density ($k_j$). Hence, we assume a fixed value of $w=14$ for our analysis.

    
    

\section{Data}
The traffic data used in this study for model training and evaluation is obtained from the California Department of Transportation's Performance Measurement System (PeMS), which is widely used in similar research \cite{Lv2014, huang2014, mallick2021}. The PeMS captures real-time traffic data from sensors along the freeway and ramps at 30 second intervals, which are aggregated every 5 minutes. 
The dataset used for model training and evaluation consists of flow, occupancy, and speed, collected from a series of vehicle detection sensors (VDS). 

\begin{figure}[!htb]
\centering
\includegraphics[width=1\columnwidth]{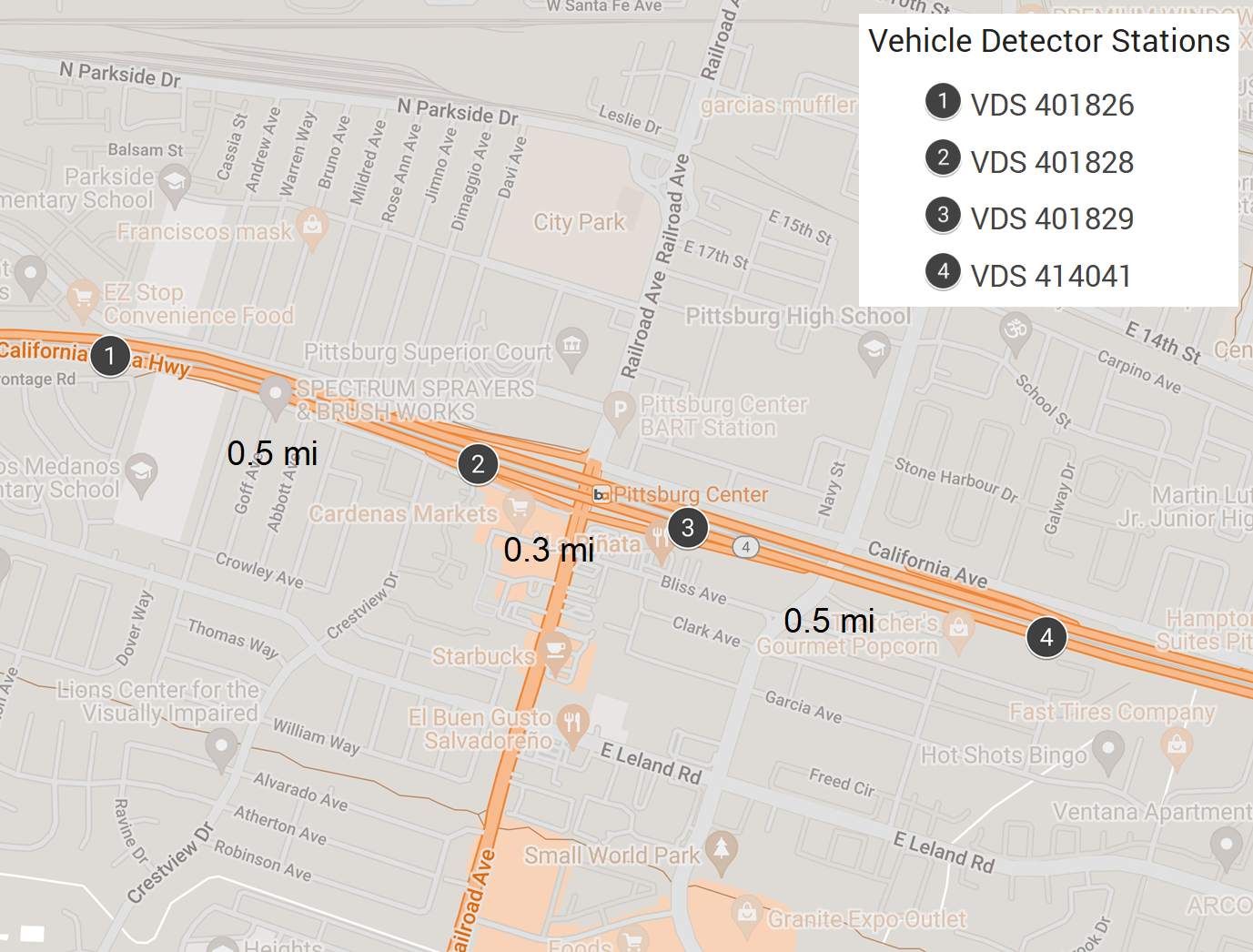}
\caption{Station configurations in Dataset 1 -- California SR04 Delta Highway}\label{fig:map1}
\end{figure}

\begin{figure}[!htb]
\centering
\includegraphics[width=1\columnwidth]{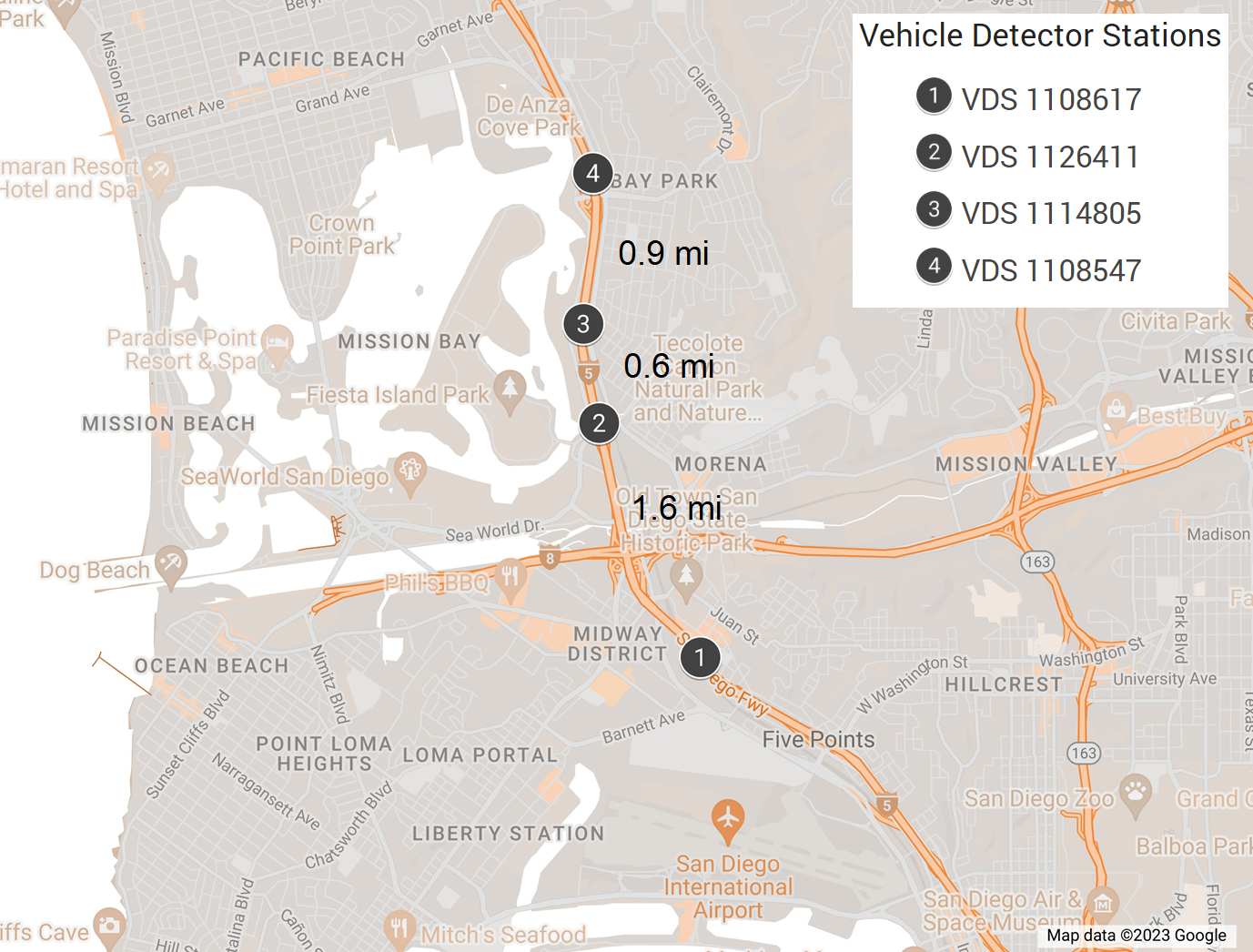}
\caption{Station configurations in Dataset 2 -- California Interstate-05 NB}\label{fig:map2}
\end{figure}

\begin{figure}[!htb]
\centering
\includegraphics[width=1\columnwidth]{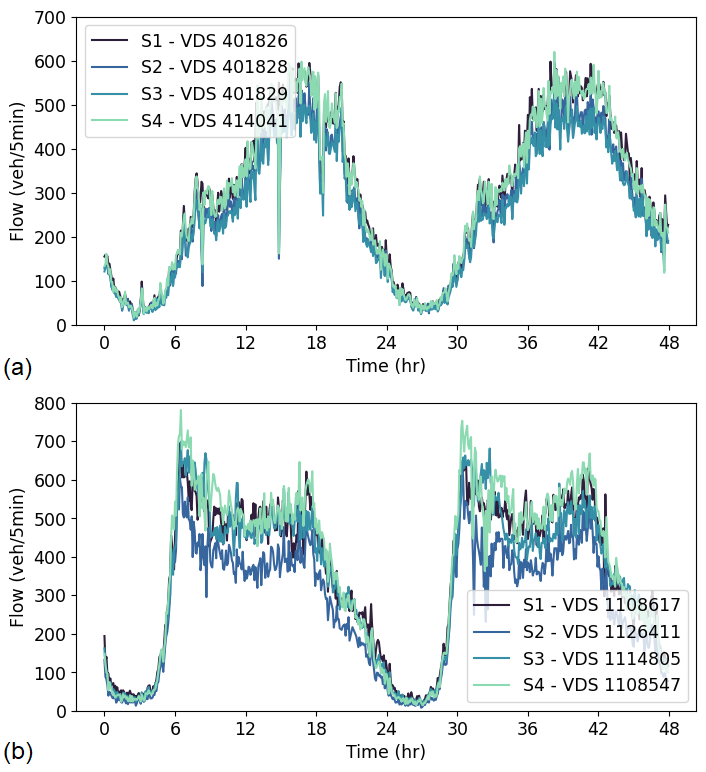}
\caption{Freeway data considered (a) California SR04 (b) I-05 NB}\label{fig:data}
\end{figure}

For this work, we use two datasets - (1) Dataset 1: California SR04 Delta Highway, collected over a period of 56 consecutive days from July 1, 2021 to August 25, 2021 including weekdays and weekends, and (2) Dataset 2:  California Interstate-05 NB in District 11, collected for one year in 2019. In Dataset 1, the distances between consecutive stations are 0.5 mi, 0.3 mi, and 0.5 mi, respectively. On the other hand, for Dataset 2, the distances between consecutive stations are 1.6 mi, 0.6 mi, and 0.9 mi. These datasets provide an understanding of how the model can scale spatially by considering the proximity and distances between monitoring stations along the freeways. 
Figures~\ref{fig:map1} and \ref{fig:map2} depicts the map of the station configurations, illustrating the locations and distances between monitoring stations for both Dataset 1 (California SR04 Delta Highway) and Dataset 2 (California Interstate-05 NB in District 11). In addition, Figure~\ref{fig:data} displays the temporal patterns of flows at various stations in both Dataset 1 and Dataset 2.

It is important to note that on- and off-ramp traffic volumes were not always available. Hence, the vehicle inflows and outflows associated with on-ramps and off-ramps could not be adjusted to maintain vehicle conservation along the freeway. As a result, the analysis and modeling in this study focus solely on the traffic data collected from the detection sensors along the main freeway lanes, without accounting for the ramp movements.

\section{Results and discussions}

The goal is to predict traffic flow 5 minutes into the future at a station where data is not available by utilizing input data from neighboring detectors for the the last 50 minutes for Dataset 1 and 100 minutes for Dataset 2. Sensitivity analysis on the performance for longer prediction time periods (from 5 minutes to 25 minutes) are discussed later.

\subsection{Prediction scenario}
To evaluate and compare the performance of the models, different scenarios are considered based on the relative positions of the target and transfer stations. This relative positioning determines which Newell's transformation is applied to the target and transfer locations. Table~\ref{tab:predreg} presents these scenarios, labeled A1, A2, B1, B2, C1, C2, D1, and D2, where the model's performance is assessed at both the target (where it has been trained) and the transfer locations without further retraining. It is assumed that up-to-date data is available at the two source locations, while no data is available at the target or transfer locations for prediction purposes. For each scenario, a regular model and three physics-modified models are considered. 

Different scenarios are considered to better understand when the proposed models can improve predictions. For instance, let us consider a scenario where the model is trained using traffic information from an upstream and downstream station to predict flows at an intermediate target location (Scenario A1 and A2). In this case, the physics-modified feature inputs consist of Newell's uncongested flow estimate (from the upstream station) and either an uncongested or congested flow estimate (from the downstream station). If the same model is applied to another intermediate location within the bounds of the upstream and downstream stations, the model will still receive similar flow estimates as inputs. However, in other scenarios (e.g., D1 or D2), the models are trained to predict flow at an upstream (or downstream) location relative to both stations but are transferred to predict flow at a downstream (or upstream) location. Since the input features in the target and transfer domains differ significantly, it is important to investigate the model's performance under these circumstances as well.

It is worth noting that Case B always utilizes free-flow features both in the target and transfer locations, since both stations are upstream of target and transfer. Therefore, we have performance recorded only for the `Physics FF' method in this case. Additionally, a hybrid model could not be employed for Case D since the dimensions of the input features in the target and transfer locations would be different, making it challenging to maintain consistency in the model architecture.

\begin{table}[!ht]
\caption{Prediction performance scenario}\label{tab:predreg}	
\begin{center}
{
\begin{tabular}{lllll}
\hline
{Scenario} & {S1} & {S2} & {S3} & {S4} \\ \hline
A1 & \textbf{Source 1} & Target & \textit{Transfer} & \textbf{Source 2} \\
A2 & \textbf{Source 1} & \textit{Transfer} & Target & \textbf{Source 2} \\
B1 & \textbf{Source 1}  & \textbf{Source 2} & Target & \textit{Transfer} \\
B2 & \textbf{Source 1}  & \textbf{Source 2} & \textit{Transfer} & Target \\
C1 & \textit{Transfer} &  Target & \textbf{Source 1} & \textbf{Source 2} \\
C2 & Target & \textit{Transfer}  & \textbf{Source 1} & \textbf{Source 2} \\
D1 & Target & \textbf{Source 1}  & \textbf{Source 2} & \textit{Transfer} \\
D2 & \textit{Transfer} & \textbf{Source 1}  & \textbf{Source 2} & Target \\
\hline
\end{tabular}%
}
\end{center}
\end{table}

\subsection{Evaluation metrics}
We evaluate the model performances for both single- and multi-step prediction horizons, by comparing the prediction mean with the corresponding true flow values using three metrics: root mean squared error (RMSE), mean absolute percentage error (MAPE) and $R^2$ as defined below. 

\begin{equation} 
\mathrm{RMSE}={\sqrt{\dfrac{1}{n}\sum_{i=1}^{N} \left [y_{i}-\hat {y_{i}}\right]^{2}} }\label{eqn:rmse}
\end{equation}

\begin{equation} 
\mathrm{MAPE}=\dfrac{100\%}{N}\sum_{i=1}^{N} \lvert{ \dfrac{y_{i}-\hat {y_{i}}}{y_{i}} }\rvert \label{eqn:mape}
\end{equation}

\begin{equation} 
\mathrm{R^2}=1-\dfrac{\sum_{i=1}^{N} \left [y_{i}-\hat {y_{i}}\right]^{2}}{\sum_{i=1}^{N} \left [y_{i}-\bar {y_{i}}\right]^{2}} \label{eqn:mae}
\end{equation}
where $y_i$ represents the true value of the observation $i$, $\hat{y_i}$ is the predicted value of $y_i$ for $i=1,2,\dots T$. Both RMSE and mean absolute error (MAE) measures the error of the predictions, however, RMSE is often preferred due to its higher penalization to outliers compared to MAE, where all errors are weighed equally. On the other hand, $\mathrm{R^2}$ measures the performance of the model in terms of proportion of the variance in data that could be explained by the regression model. 

\subsection{Model training}
In this study, two different model architectures are used due to variations in the volume of the two datasets. 
For Dataset 1, the model architecture comprises a 2-dimensional convolutional layer with 12 filters and a kernel size of (3, 2), activated by the rectified linear unit (ReLU) activation function. The outputs from the convolutional layer are then flattened and reshaped appropriately to be fed into two LSTM layers with 10 and 6 units respectively. Finally, a single-unit dense layer is utilized to predict flows. 
On the other hand, for Dataset 2, the model architecture consists of a 2-dimensional convolutional layer with 16 filters and a kernel size of (3, 2), activated by the rectified linear unit (ReLU) activation function. This is followed by two LSTM layers with 10 and 6 units, respectively. The outputs from the LSTM layers are then passed through a dense layer with 6 units, activated by the ReLU activation function. Finally, the prediction layer is added to generate the traffic flow predictions for the specified future time point. 

To ensure generalizability and prevent overfitting, the models are trained, validated, and tested using three distinct sets. The dataset is partitioned into three parts: 60\% for model training, 15\% for validation, and 25\% for testing. 
The model parameters are tuned throughout the training process based on their performance on the validation set. The mean squared error (MSE) loss function is minimized during the training process, and the model with the lowest validation error is selected as the final model.
For the models trained on Dataset 1, a batch size of 10 and a temporal lag of 10 are used. In contrast, the models trained on Dataset 2 uses a batch size of 20 and a temporal lag of 20. The Adadelta \cite{adadelta} optimizer is employed with a learning rate of 0.10, a rho value of 0.95, and an epsilon value of 1e-7 to train the models. 

\subsection{Dataset 1: California SR04}
Prediction performances of the models trained on Dataset 1 are provided in Tables~\ref{tab:small_perf1}, \ref{tab:small_perf2} and \ref{tab:small_perf3} corresponding to different performance metrics -- RMSE, $R^2$ and MAPE.

\begin{table*}[!ht]
\caption{Comparison of RMSE between regular and physics-modified models trained on Dataset 1: California SR04}\label{tab:small_perf1}	
\begin{center}
\begin{tabular}{llllll}
\hline
Case & Location & Regular & Physics FC & Physics FF & Hybrid \\ \hline
\multirow{2}{*}{A1} & Target & 27.4609 & 27.4523 & 27.4251 & 27.3052 \\ \cline{2-6} 
 & Transfer & 24.1859 & 24.0482 & 24.2259 & 24.1502 \\ \hline
\multirow{2}{*}{A2} & Target & 23.9148 & 23.9316 & 23.9943 & 23.8019 \\ \cline{2-6} 
 & Transfer & 27.5623 & 27.7183 & 27.6590 & 27.4935 \\ \hline
\multirow{2}{*}{B1} & Target & 23.9541 & - & 23.9417 & - \\ \cline{2-6} 
 & Transfer & 26.6676 & - & 26.6647 & - \\ \hline
\multirow{2}{*}{B2} & Target & 26.0435 & - & 25.9842 & - \\ \cline{2-6} 
 & Transfer & 24.2526 & - & 24.4648 & - \\ \hline
\multirow{2}{*}{C1} & Target & 27.8632 & 27.8772 & 27.8294 & 27.8290 \\ \cline{2-6} 
 & Transfer & 29.3075 & 30.6976 & 29.2691 & 29.5725 \\ \hline
\multirow{2}{*}{C2} & Target & 29.0854 & 30.4303 & 29.0880 & 29.0852 \\ \cline{2-6} 
 & Transfer & 28.0220 & 28.3145 & 28.1024 & 28.0230 \\ \hline
\multirow{2}{*}{D1} & Target & 29.3008 & 29.2855 & 29.2226 & - \\ \cline{2-6} 
 & Transfer & 27.3864 & 29.4767 & 27.4675 & - \\ \hline
\multirow{2}{*}{D2} & Target & 26.5213 & 26.5085 & 26.5085 & - \\ \cline{2-6} 
 & Transfer & 29.5364 & 29.3480 & 28.5449 & - \\ \hline
\end{tabular}
\end{center}
\end{table*}

\begin{table*}[!ht]
\caption{Comparison of $R^2$ between regular and physics-modified models trained on Dataset 1: California SR04}\label{tab:small_perf2}	
\begin{center}
\begin{tabular}{llllll}
\hline
Case & Location & Regular & Physics FC & Physics FF & Hybrid \\ \hline
\multirow{2}{*}{A1} & Target & 0.9648 & 0.9648 & 0.9649 & 0.9652 \\ \cline{2-6} 
 & Transfer & 0.9705 & 0.9709 & 0.9704 & 0.9706 \\ \hline
\multirow{2}{*}{A2} & Target & 0.9712 & 0.9711 & 0.9710 & 0.9714 \\ \cline{2-6} 
 & Transfer & 0.9645 & 0.9641 & 0.9643 & 0.9647 \\ \hline
\multirow{2}{*}{B1} & Target & 0.9711 & - & 0.9711 & - \\ \cline{2-6} 
 & Transfer & 0.9739 & - & 0.9739 & - \\ \hline
\multirow{2}{*}{B2} & Target & 0.9751 & - & 0.9752 & - \\ \cline{2-6} 
 & Transfer & 0.9703 & - & 0.9698 & - \\ \hline
\multirow{2}{*}{C1} & Target & 0.9638 & 0.9637 & 0.9638 & 0.9639 \\ \cline{2-6} 
 & Transfer & 0.9692 & 0.9663 & 0.9693 & 0.9687 \\ \hline
\multirow{2}{*}{C2} & Target & 0.9697 & 0.9668 & 0.9697 & 0.9697 \\ \cline{2-6} 
 & Transfer & 0.9633 & 0.9626 & 0.9631 & 0.9633 \\ \hline
\multirow{2}{*}{D1} & Target & 0.9693 & 0.9693 & 0.9694 & - \\ \cline{2-6} 
 & Transfer & 0.9724 & 0.9681 & 0.9723 & - \\ \hline
\multirow{2}{*}{D2} & Target & 0.9741 & 0.9742 & 0.9742 & - \\ \cline{2-6} 
 & Transfer & 0.9688 & 0.9692 & 0.9708 & - \\ \hline
\end{tabular}
\end{center}
\end{table*}

\begin{table*}[!ht]
\caption{Comparison of MAPE between regular and physics-modified models trained on Dataset 1: California SR04}\label{tab:small_perf3}	
\begin{center}
\begin{tabular}{llllll}
\hline
Case & Location & Regular & Physics FC & Physics FF & Hybrid \\ \hline
\multirow{2}{*}{A1} & Target & 0.8093 & 0.7971 & 0.8065 & 0.7657 \\ \cline{2-6} 
 & Transfer & 0.7156 & 0.6964 & 0.7079 & 0.6809 \\ \hline
\multirow{2}{*}{A2} & Target & 0.7183 & 0.7008 & 0.7070 & 0.6872 \\ \cline{2-6} 
 & Transfer & 0.8201 & 0.8149 & 0.8291 & 0.7958 \\ \hline
\multirow{2}{*}{B1} & Target & 0.7136 & - & 0.7184 & - \\ \cline{2-6} 
 & Transfer & 1.0218 & - & 1.0394 & - \\ \hline
\multirow{2}{*}{B2} & Target & 0.9575 & - & 0.9838 & - \\ \cline{2-6} 
 & Transfer & 0.7440 & - & 0.7956 & - \\ \hline
\multirow{2}{*}{C1} & Target & 0.8096 & 0.8086 & 0.8139 & 0.7681 \\ \cline{2-6} 
 & Transfer & 5.0663 & 4.6473 & 5.0910 & 5.1809 \\ \hline
\multirow{2}{*}{C2} & Target & 5.0956 & 5.0455 & 5.1540 & 5.2339 \\ \cline{2-6} 
 & Transfer & 0.8543 & 0.8717 & 0.8476 & 0.8487 \\ \hline
\multirow{2}{*}{D1} & Target & 5.2534 & 4.8477 & 5.1256 & - \\ \cline{2-6} 
 & Transfer & 1.0569 & 0.9453 & 1.0630 & - \\ \hline
\multirow{2}{*}{D2} & Target & 1.0126 & 1.0126 & 1.0127 & - \\ \cline{2-6} 
 & Transfer & 4.5709 & 4.2736 & 4.3245 & - \\ \hline
\end{tabular}
\end{center}
\end{table*}

The performance results for Cases A, C, and D demonstrate that Newell's assumption of congested states prevailing downstream does not hold true, or the interference of ramp flows are too large for Newell's conservation assumptions to hold. As a result, the `Physics FC' method, which relies on this assumption, does not outperform the regular model either at the target or transfer locations in the cases where the source station is downstream of either the target or transfer location, or both.
In contrast, we find that the assumption of free-flow states performs better in the majority of these scenarios. 

The hybrid approach demonstrates superior performance in both Cases A and C compared to other approaches. In Cases A1 and A2, the feature inputs consist of three dimensions: one from the upstream station and two from the downstream stations. This configuration allows the model to leverage the free-flow characteristics from both upstream and downstream stations, as well as the congested information from downstream, thereby enhancing its prediction capabilities. 
Similarly, in Cases C1 and C2, the input feature size is expanded to four dimensions, with two dimensions from both the upstream and downstream stations. This enables the model to incorporate the combined information from both directions, taking into account the variations in traffic conditions along the freeway section. By considering the inputs from both upstream and downstream stations, the model can better capture the complex interactions and correlations between different segments of the freeway, leading to improved prediction accuracy.

\subsection{Dataset 2: Interstate-05 NB}
\begin{figure}[!htb]
\centering
\includegraphics[width=0.85\columnwidth]{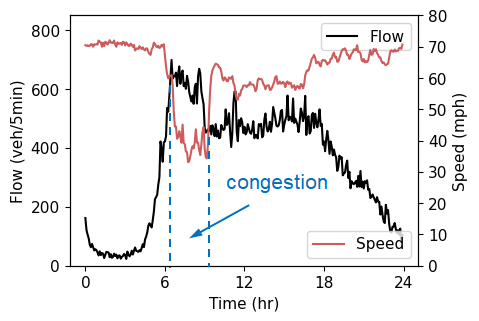}
\caption{Free-flow and congestion states for VDS 1114805 }\label{fig:congestion}
\end{figure}

Here, we present the modeling results on Dataset 2 in Table~\ref{tab:large_perf}, which displays the RMSE values. Other evaluation measures exhibit similar trends and are not included in the table for brevity. 
In this larger dataset, we also evaluate the accuracy of the model predictions separately during free flow and congestion to better understand the benefits of the proposed approaches. To distinguish between free-flow and congestion states, we divide the daily traffic into two categories based on a speed threshold:  if the speed drops below 50 mph it is assumed that that location is in congestion (as shown in blue dashed lines), See Figure~\ref{fig:congestion}.

\begin{table*}[!ht]
\caption{Comparison of RMSE between regular and physics-modified models trained on Dataset 2: Interstate-05 NB}\label{tab:large_perf}	
\begin{center}
\begin{tabular}{lllllll}
\hline
Case & State & Location & Regular & Physics FC & Physics FF & Hybrid \\ \hline
\multirow{6}{*}{A1} & \multirow{2}{*}{Combined} & Target & 27.6386 & 27.7352 & 27.6706 & 27.4902 \\ \cline{3-7} 
 &  & Transfer & 41.4230 & 41.5126 & 42.1914 & 41.4253 \\ \cline{2-7} 
 & \multirow{2}{*}{Free-flow} & Target & 24.7315 & 24.7775 & 24.9809 & 24.5950 \\ \cline{3-7} 
 &  & Transfer & 37.2959 & 37.7488 & 38.0739 & 37.4430 \\ \cline{2-7} 
 & \multirow{2}{*}{Congestion} & Target & 38.6853 & 38.9359 & 38.0415 & 38.4616 \\ \cline{3-7} 
 &  & Transfer & 57.2159 & 56.1263 & 58.0150 & 56.7572 \\ \hline
\multirow{6}{*}{A2} & \multirow{2}{*}{Combined} & Target & 46.5795 & 48.0333 & 47.0279 & 45.6740 \\ \cline{3-7} 
 &  & Transfer & 38.2043 & 39.8507 & 38.1397 & 37.5140 \\ \cline{2-7} 
 & \multirow{2}{*}{Free-flow} & Target & 47.1252 & 48.8942 & 47.5663 & 46.3539 \\ \cline{3-7} 
 &  & Transfer & 32.0716 & 33.4739 & 32.1922 & 31.6403 \\ \cline{2-7} 
 & \multirow{2}{*}{Congestion} & Target & 44.4094 & 44.3008 & 44.8736 & 42.8027 \\ \cline{3-7} 
 &  & Transfer & 59.4009 & 61.9251 & 58.8517 & 57.9369 \\ \hline
\multirow{6}{*}{B1} & \multirow{2}{*}{Combined} & Target & 38.5346 & - & 38.6557 & - \\ \cline{3-7} 
 &  & Transfer & 43.0122 & - & 42.2568 & - \\ \cline{2-7} 
 & \multirow{2}{*}{Free-flow} & Target & 38.2690 & - & 38.4363 & - \\ \cline{3-7} 
 &  & Transfer & 39.9212 & - & 39.0535 & - \\ \cline{2-7} 
 & \multirow{2}{*}{Congestion} & Target & 39.9460 & - & 39.8573 & - \\ \cline{3-7} 
 &  & Transfer & 55.7046 & - & 55.2978 & - \\ \hline
\multirow{6}{*}{B2} & \multirow{2}{*}{Combined} & Target & 36.5822 & - & 36.6687 & - \\ \cline{3-7} 
 &  & Transfer & 32.0468 &-  & 31.1714 &  -\\ \cline{2-7} 
 & \multirow{2}{*}{Free-flow} & Target & 35.4826 & - & 35.5825 & - \\ \cline{3-7} 
 &  & Transfer & 29.7955 & - & 28.9267 & - \\ \cline{2-7} 
 & \multirow{2}{*}{Congestion} & Target & 41.9924 & - & 42.0334 & - \\ \cline{3-7} 
 &  & Transfer & 41.3182 & - & 40.3653 & - \\ \hline
 \multirow{6}{*}{C1} & \multirow{2}{*}{Combined} & Target & 32.7486 & 32.1711 & 32.9541 & 33.0664 \\ \cline{3-7} 
 &  & Transfer & 41.3909 & 45.6134 & 42.0968 & 44.0113 \\ \cline{2-7} 
 & \multirow{2}{*}{Free-flow} & Target & 29.8942 & 29.4090 & 30.3760 & 30.6260 \\ \cline{3-7} 
 &  & Transfer & 38.6701 & 42.6464 & 39.2933 & 39.8573 \\ \cline{2-7} 
 & \multirow{2}{*}{Congestion} & Target & 44.0564 & 43.1335 & 43.3757 & 43.0108 \\ \cline{3-7} 
 &  & Transfer & 52.6719 & 57.8503 & 53.1065 & 60.1006 \\ \hline
\multirow{6}{*}{C2} & \multirow{2}{*}{Combined} & Target & 37.1587 & 39.6226 & 37.9617 & 36.9906 \\ \cline{3-7} 
 &  & Transfer & 36.0370 & 36.9692 & 36.3954 & 35.7617 \\ \cline{2-7} 
 & \multirow{2}{*}{Free-flow} & Target & 36.5649 & 39.0817 & 37.6206 & 36.7317 \\ \cline{3-7} 
 &  & Transfer & 31.5107 & 32.5991 & 32.4174 & 31.6026 \\ \cline{2-7} 
 & \multirow{2}{*}{Congestion} & Target & 40.1009 & 42.3659 & 39.7900 & 38.4324 \\ \cline{3-7} 
 &  & Transfer & 52.5501 & 53.1936 & 51.3670 & 51.2351 \\ \hline
\multirow{6}{*}{D1} & \multirow{2}{*}{Combined} & Target & 33.5547 & 35.1328 & 33.4300 & - \\ \cline{3-7} 
 &  & Transfer & 46.7479 & 50.9093 & 46.8157 & - \\ \cline{2-7} 
 & \multirow{2}{*}{Free-flow} & Target & 33.0796 & 34.7593 & 32.8865 & - \\ \cline{3-7} 
 &  & Transfer & 41.2508 & 42.4285 & 41.3332 & - \\ \cline{2-7} 
 & \multirow{2}{*}{Congestion} & Target & 35.8799 & 37.0699 & 36.0348 & - \\ \cline{3-7} 
 &  & Transfer & 67.3285 & 79.9866 & 67.3495 & - \\ \hline
\multirow{6}{*}{D2} & \multirow{2}{*}{Combined} & Target & 35.8765 & 36.1171 & 36.1171 & - \\ \cline{3-7} 
 &  & Transfer & 40.5691 & 45.7459 & 37.9290 & - \\ \cline{2-7} 
 & \multirow{2}{*}{Free-flow} & Target & 34.7978 & 35.0323 & 35.0323 & - \\ \cline{3-7} 
 &  & Transfer & 36.6836 & 40.0097 & 34.5566 & - \\ \cline{2-7} 
 & \multirow{2}{*}{Congestion} & Target & 41.1896 & 41.4576 & 41.4576 & - \\ \cline{3-7} 
 &  & Transfer & 55.5357 & 66.6779 & 51.0994 & - \\ \hline

\end{tabular}
\end{center}
\end{table*}

The results for Cases A1 and A2, where both the target and transfer locations are in between the source detector stations, again indicate that the use of Newell's modification for congestion conditions does not improve predictions, likely due to lack of congestion or interference of traffic dynamics due to ramp flows. This finding aligns with the observations from Dataset 1. Interestingly, assuming free-flow conditions leads to only minor improvements in some cases. 
However, when we employ the hybrid approach that combines information from both upstream (free-flow) and downstream (congested and uncongested) stations, we observe significant improvements in model performance. This hybrid approach outperforms both the regular model and other versions of the physics-based approaches. It is important to note that during the model training, the optimization is still based on the combined mean squared error (MSE) loss, meaning that the model does not individually optimize for losses during free-flow and congestion. The superior performance of the hybrid approach highlights the effectiveness of incorporating both free-flow and congestion information to enhance the model's predictive capabilities. 

In Case C1, we observe that Newell's assumption of congestion prevailing downstream (Physics FC) works well at the target location (S2), where only congestion characteristics from downstream locations are used during model training. However, when this assumption is transferred to S1, it does not perform well. In contrast, the assumption of free-flow conditions (Physics FF) shows poor performance at the target location but demonstrates better performance at the transfer location in Case C1. Surprisingly, the hybrid approach also shows poor overall performance in both the target and transfer locations. But it still provides better performance during congestion at the target location.

\begin{figure*}[!htb]
\centering
\includegraphics[width=0.85\textwidth]{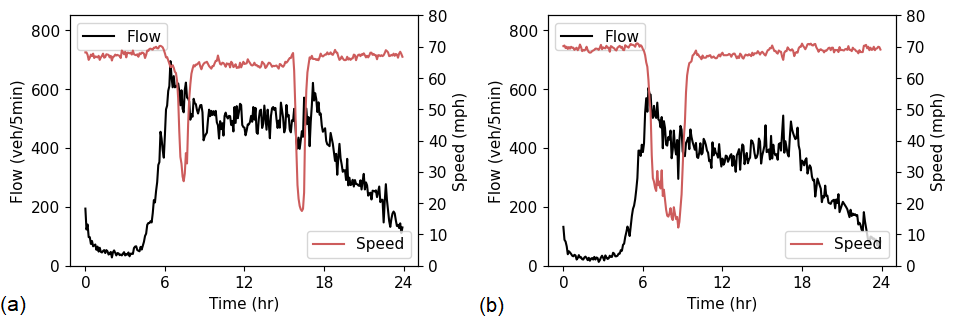}
\caption{Flow vs. speed profiles for (a) VDS 1108617 and (b) VDS 1126411}\label{fig:congestion_compare}
\end{figure*}

The discrepancy in performance can be attributed to location-specific differences, such as variations in traffic patterns, congestion levels, or spatial configurations of the freeway section. In this case, the target station for Case C1 (S2) is located 0.6 mi and 1.5 mi away from the source stations, while the transfer location for Case C1 (S1) is located 2.2 mi and 3.1 mi away. These differences in spatial distances can significantly impact the traffic dynamics and patterns observed at each location. 
To further support this observation, we examine the flow vs. speed trends for both S2 (target in Case C1) and S1 (transfer in Case C1) as depicted in Figure~\ref{fig:congestion_compare}. When comparing these trends, we find that S1 operates predominantly in free-flow conditions with only short-lived congested states. This aligns with the better transferability of the FF model, which leverages the free-flow characteristics, as opposed to the hybrid model that combines both free-flow and congested information. 

In Case C2, we observe a different pattern compared to Case C1. The hybrid approach performs better than both the regular model and the `Physics FF' approach in both the target and transfer locations. The `Physics FF' model also demonstrates good performance, consistent with the assumption that the target station mostly operates in free-flow conditions. However, the hybrid approach excels by accurately accounting for the dependencies between the source and target station.

In Cases B1 and B2, we observe that while the regular model performs better at the target location, its transferability is compromised, indicating that it does not generalize well to different locations. On the other hand, the physics-based FF model demonstrates better transferability in these cases. This suggests that the `Physics FF' model captures the underlying dynamics of traffic more effectively, allowing it to perform well even at locations it has not been explicitly trained on. 

In Cases D1 and D2, we consistently observe that the `Physics FF' model outperforms the `Physics FC' model, indicating that free-flow characteristics play a significant role in these scenarios. However, it is important to note that there are some instances where the regular model shows minor advantages. 

These inconsistencies in performance could be attributed to several factors. Firstly, the spatial distances between stations in this dataset are larger, which may violate the assumption of homogeneous traffic parameters over the entire section. Secondly, the presence of major interchanges and ramps between the mainline detectors can significantly impact traffic patterns. When ramps experience high volumes of vehicles trying to enter or exit the highway, they can disrupt the smooth flow of traffic and lead to localized congestion or bottlenecks on the mainline. For example, congestion on exit ramps can lead to queue spillbacks, causing disruptions in traffic flow on the mainline.
These disruptions can result in variations in traffic conditions that may not be accurately captured by the models.

Overall, while the `Physics FF' and hybrid models (where applicable) generally outperform the regular model and `Physics FC' in terms of transferability, the inconsistencies in trends compared to Dataset 1 highlight the influence of spatial distances, non-homogeneous traffic parameters, and the presence of interchanges and ramps on the accuracy of the models. These factors should be carefully considered when applying and interpreting the modeling results in such complex freeway environments.

\subsection{Time sensitivity}
Performing a sensitivity analysis along the time dimension is crucial to assess the robustness of the model when predicting traffic flows further into the future. In this case, both the regular model and the physics-based hybrid model were trained on Dataset 1 for Case A. The models were evaluated to predict 25 minutes into the future using a temporal history of 50 minutes, which consists of 10 time steps. 
The analysis reveals that the prediction errors for both models increase as the prediction horizon becomes longer. This is expected, as longer prediction horizons introduce more uncertainties and make it more challenging to accurately forecast traffic flows.
However, it is observed that the physics-based hybrid model consistently outperforms the regular model at all prediction horizons. The performance gap between the two models tends to widen as the prediction horizon increases, indicating the superiority of the physics-based approach in capturing the dynamics of traffic and improving the accuracy of predictions.
Furthermore, the physics-based hybrid model demonstrates relatively higher benefits at the transfer location compared to the target location. This suggests that incorporating physics-based principles and combining information from both upstream and downstream stations allows for a more comprehensive representation of the traffic conditions and enhances the model's predictive capabilities.

\begin{figure}[!htb]
\centering
\includegraphics[width=0.85\columnwidth]{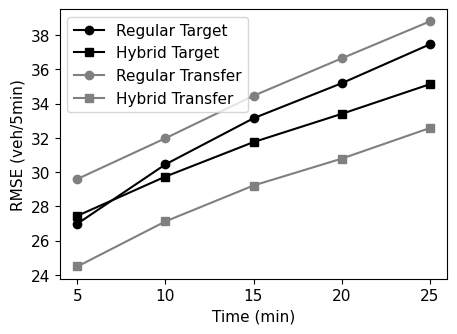}
\caption{Sensitivity analysis with respect to time}\label{fig:time_sens}
\end{figure}

\section{Conclusions}
Real-time traffic prediction on freeways plays a crucial role in the effective functioning of Intelligent Transportation Systems. Extensive research has been conducted to enhance the accuracy of traffic prediction models, including statistical parametric and non-parametric approaches that provide improved flexibility in capturing the spatial and temporal aspects of traffic patterns. However, despite these advancements, state-of-the-art spatio-temporal models such as CNN-LSTM often learn case-specific features that lack generalizability. These models uncover only limited information about the underlying traffic dynamics, thereby limiting their usability to specific training conditions.

In this study, we propose a feature transformation approach for traffic flow prediction models based on traffic flow theory. The proposed model demonstrates the ability to transfer its learned features to different settings, thereby enhancing its generalizability. The results indicate that the physics-based model can learn more universal features by mapping true traffic flows using estimates of both congested and uncongested flows. When compared to the regular model, the physics-based models exhibits improved prediction performance at both the target and transfer locations across various prediction scenarios. This improvement can be attributed to the inherent ability of the proposed model's feature inputs to account for spatial shifts, making them more transferable.
Our analysis reveals that physics-based FC models trained using congested state estimators from downstream locations do not perform well. This lack of performance could be attributed to the absence of congested states downstream or the significant interference of ramp flows, which violate Newell's conservation assumptions. On the other hand, FF and hybrid models generally outperform the regular model and FC model in terms of transferability. Minor discrepancies in trends for larger datasets are observed, likely due to location-specific attributes. Additionally, the physics-based hybrid model consistently outperforms the regular model across all prediction horizons in a multi-step prediction setting.
However, due to limitations in the available data, we were unable to conduct spatial sensitivity analysis, highlighting the need for further research using simulated data.

\bibliography{manuscript.bib}
\end{document}